\documentclass{article}

\usepackage{arxiv}

\usepackage[utf8]{inputenc} 
\usepackage[T1]{fontenc}    
\usepackage{hyperref}       
\usepackage{url}            
\usepackage{booktabs}       
\usepackage{amsfonts}       
\usepackage{nicefrac}       
\usepackage{microtype}      
\usepackage{lipsum}
\usepackage{graphicx}
\usepackage{lineno,hyperref}
\usepackage{caption}
\usepackage{subcaption}
\usepackage{booktabs,tabularx}

\usepackage{algorithm}
\usepackage{amsmath}
\usepackage[noend]{algpseudocode}

\algnewcommand\algorithmicforeach{\textbf{for each}}
\algdef{S}[FOR]{ForEach}[1]{\algorithmicforeach\ #1\ \algorithmicdo}
\DeclareMathOperator*{\argmax}{arg\,max}

\title{Multi-Camera Vehicle Counting Using Edge-AI}

\author{
  Luca Ciampi\thanks{Institute of Information Science and Technologies - National Research Council - Pisa, Italy. \newline Corresponding author: luca.ciampi@isti.cnr.it} 
   \And
 Claudio Gennaro$^{*}$
   \And
 Fabio Carrara$^{*}$
     \And
 Fabrizio Falchi$^{*}$
     \And
 Claudio Vairo$^{*}$
     \And
 Giuseppe Amato$^{*}$
}

\begin{document}
\maketitle

\begin{abstract}
This paper presents a novel solution to automatically count vehicles in a parking lot using images captured by smart cameras. Unlike most of the literature on this task, which focuses on the analysis of \textit{single} images, this paper proposes the use of multiple visual sources to monitor a wider parking area from different perspectives. The proposed multi-camera system is capable of automatically estimate the number of cars present in the \textit{entire} parking lot directly on board the edge devices. It comprises an on-device deep learning-based detector that locates and counts the vehicles from the captured images and a decentralized geometric-based approach that can analyze the inter-camera shared areas and merge the data acquired by all the devices. We conduct the experimental evaluation on an extended version of the \textit{CNRPark-EXT} dataset, a collection of images taken from the parking lot on the campus of the National Research Council (CNR) in Pisa, Italy. We show that our system is robust and takes advantage of the redundant information deriving from the different cameras, improving the overall performance without requiring any extra geometrical information of the monitored scene.
\end{abstract}


\section{Introduction}
\label{sec:intro}
Traffic-related issues are constantly increasing, and tomorrow's cities cannot be considered intelligent if they do not enable smart mobility. Smart mobility applications, such as smart parking and road traffic management, are nowadays widely employed worldwide, making our cities more livable and bringing benefits to the cities and, consequently, to our lives.

Images are perhaps the best sensing modality to perceive and assess the flow of vehicles in large areas. Like no other sensing mechanism, city camera networks can monitor large areas while simultaneously providing visual data to AI systems to extract relevant information from this deluge of data. However, this application is often hampered by the massive flow of data that must be sent to central servers or the cloud for processing.
On the other hand, edge computing is a recent paradigm that promotes the decentralization of data processing to the border, i.e., where the data are generated, thus reducing the traffic on the network and the pressure on central servers. No wonder that the combination of recent Computer Vision techniques like the deep learning-based ones and the edge computing paradigm is an emerging trend, although they must face the limited computational resources on the disposable edge devices.

\begin{figure}[htbp]
\centerline{\includegraphics[width=.60\textwidth]{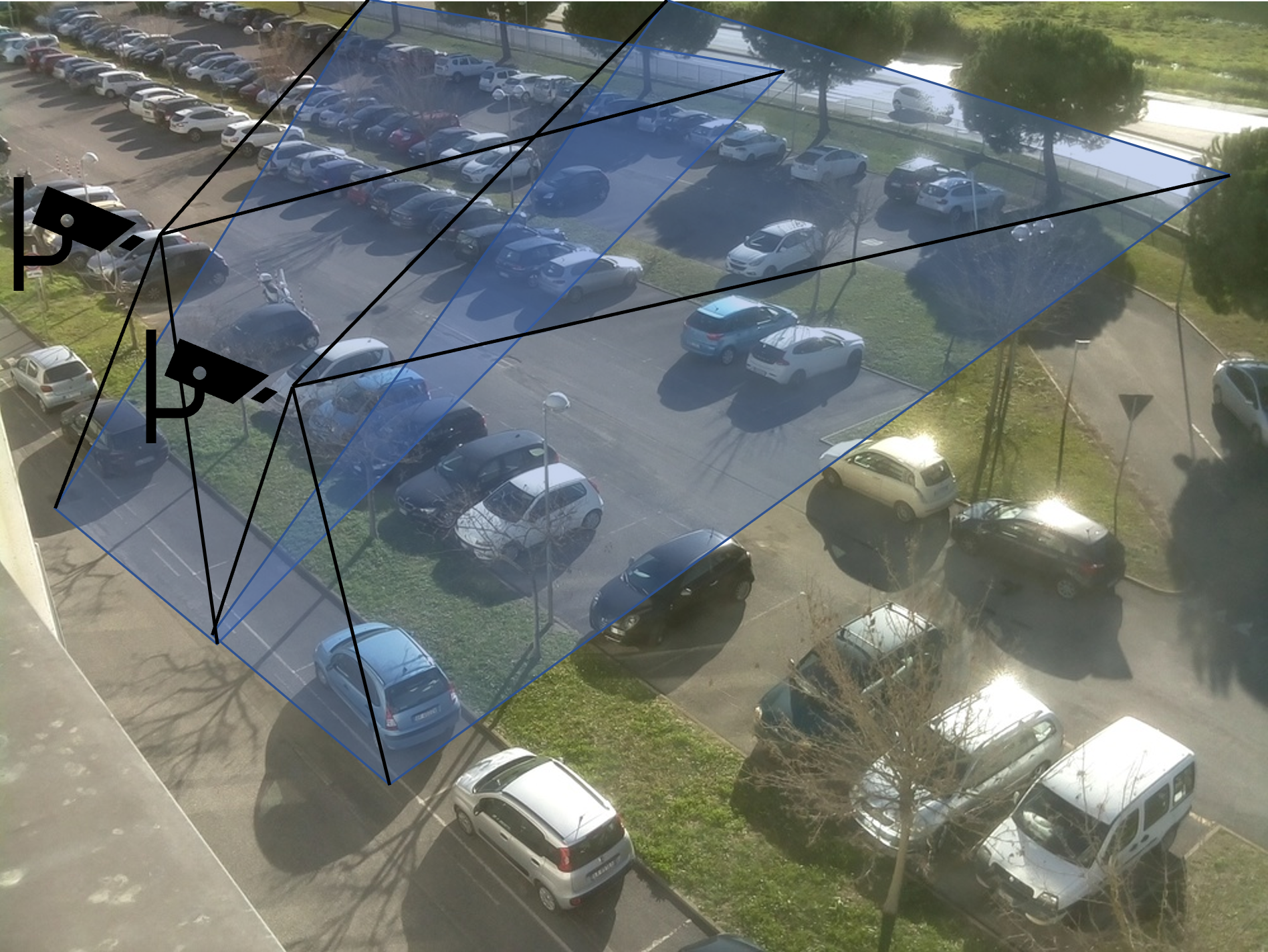}}
\caption{An example of two cameras monitoring the same parking area with partially overlapping fields of views.
This redundancy provides robustness and fault-tolerance but also raises the problem of aggregating knowledge extracted from the individual cameras.
}
\label{fig:multi_camera_example}
\end{figure}

In this work, we propose a novel solution to automatically estimate the number of vehicles present in a parking lot using images captured by smart cameras. This counting task is challenging as the process of understanding the captured images faces many problems, such as shadows, light variation, weather conditions, and inter-object occlusions. Most of the existing works concerning the vehicles counting task focus on the analysis of \textit{single} images. However, in many real-world scenarios, one can benefit from using multiple cameras to monitor the same parking lot from different perspectives and viewpoints. Furthermore, multiple neighboring cameras can also be helpful to cover a wider area. At the same time, such an approach introduces issues related to merging the knowledge extracted from the single cameras with partially overlapping fields of views (FOVs), as shown in Figure \ref{fig:multi_camera_example}.

In this paper, we introduce a multi-camera system that combines a CNN-based technique, which can locate and count vehicles present in images belonging to individual cameras, along with a decentralized geometry-based approach that is responsible for aggregating the data gathered from all the devices and estimating the number of cars present in the \textit{entire} parking lot. Our solution performs the task directly on the edge devices (i.e., the smart cameras) without using a central server or cloud, consequently reducing the communication overhead. The total count is built exploiting the partial results computed in parallel by the single cameras and propagated through messages. Hence, our system scales better when the number of monitored parking spaces increases. 
Moreover, our solution does not require any extra information about the monitored parking area, such as the location of the parking spaces, nor any geometric information about the camera positions in the parking lot. In short, it is a flexible and ready-to-use solution that allows a simple ``plug-and-play'' insertion of new cameras into the system.

To train the CNN and validate our solution, we employ the \textit{CNRPark-EXT} dataset \cite{amato2017deep}, a collection of images taken from the parking lot on the campus of the National Research Council (CNR) in Pisa, Italy. The pictures are acquired by multiple cameras having partially overlapping fields of views and describing challenging scenarios, with different perspectives, illuminations, weather conditions, and many occlusions. Since the annotations of this dataset concern single images, we extended it by relabeling a part of it to be consistent with our algorithm that instead considers the entire parking area. We conduct extensive experiments testing the generalization capabilities of the CNN-based technique responsible for detecting vehicles in single images and the effectiveness of our multi-camera algorithm, demonstrating that our system is robust and benefits from the redundant information deriving from the different cameras improving the overall performance.

To summarizing, the main contributions of this work are the followings:
\begin{itemize}
\item We introduce a novel multi-camera system able to automatically estimate the number of cars present in the \textit{entire} monitored parking area. It runs directly on the edge devices and combines a deep learning-based detector together with a decentralized technique that exploits the geometry of the captured images.
\item We specifically extend the \textit{CNRPark-EXT} dataset \cite{amato2017deep}, a collection of images acquired by multiple cameras having partially overlapping fields of views and describing various parking lots. We manually re-label a subset of it, making it suitable with our considered scenario in which we consider the whole parking area.
\item We conduct experiments showing that our system is robust, flexible, and can benefit from redundant information coming from different cameras while improving overall performance.
\end{itemize}

We organize the rest of the paper as follows. Section \ref{sec:related_work} reports other works present in the literature related to our topic. Section \ref{sec:proposed_solution} describes our multi-camera counting algorithm. Section \ref{sec:exp_setup} states the experimental setup, describing the dataset, the metrics, and the implementation details. Section \ref{sec:experiments} presents and discusses the experiments and the obtained results. Finally, Section \ref{sec:conclusions} concludes the paper with some insights on future directions.

\section{Related Work}
\label{sec:related_work}
In this section, we overview some works related to our, organizing them into two categories. The first one concerning the counting task, while the second one regarding multi-camera parking lot monitoring systems.

\subsection{The counting task}
The counting task estimates the number of object instances in still images or video frames \cite{lempitsky2010learning}. This topic has recently attracted much attention due to its inter-disciplinary and widespread applicability and its paramount importance for many real-world applications. Examples include counting bacterial cells from microscopic images \cite{xie2018microscopy}, estimate the number of people present at an event \cite{boominathan2016crowdnet}, counting animals in ecological surveys to monitor the population of a specific region \cite{arteta2016counting} and evaluate the number of vehicles on a highway or in a car park \cite{amato2019counting}.

In the last years, several machine learning-based solutions (especially supervised) have been suggested. Following the taxonomy adopted in \cite{sindagi2018survey}, we can broadly classify existing counting approaches into two categories: counting by regression and counting by detection. Counting by \textit{regression} is a supervised method that tries to establish a direct mapping (linear or not) from the image features to the number of objects present in the scene or a corresponding density map (i.e., a continuous-valued function), skipping the challenging task of detecting instances of the objects \cite{zhang2016single, zhang2017understanding, onoro2016towards, DBLP:conf/ecai/CiampiSCGA20, DBLP:conf/visapp/CiampiSCGA21}. Counting by \textit{detection} is, instead, a supervised approach where we localize instances of the objects, and then we count them \cite{amato2018wireless, ciampi2018counting}. While regression-based techniques work very well in a scenario where the objects are extremely overlapped, and the single instances are not well defined due to inter-class and intra-class occlusions, they perform poorly in images with a large perspective and oversized objects.

In this work, we estimate the number of vehicles present in a park area from images collected by smart cameras having large perspectives. The cars close to the cameras are much larger than the ones far away from them. Therefore, we employ a detection-based method. Most of the existing counting solutions do not directly deal with edge computing devices and the consequent constraints due to the limited available computing resources. They use deep learning-based approaches that typically require the use of a GPU and that are computationally expensive. Moreover, they consider the images as single entities. They do not account for the possible benefits of monitoring the same lots from different perspectives or covering a wider parking area with multiple cameras. Instead, our solution runs directly on the edge devices connected to each other and can estimate the number of vehicles present in the entire parking lot.

\subsection{Multi-camera parking lot monitoring}
Parking lot monitor using visual data is not new, and other works already tackled it in the literature. In \cite{amato2016car,amato2017deep}, the same authors of the \textit{CNRPark-EXT} dataset presented a deep learning-based system for parking lot occupancy detection that can run in a Raspberry Pi directly on-board a smart camera. In \cite{nieto2018automatic}, authors directly dealt with the issues deriving from the adoption of a multi-camera system. In particular, they applied a homography to project the detected vehicles from the plane of each camera to a common plane, where they performed a perspective correction to correct matching between the vehicle detections and the parking spots. Also, the authors in \cite{vitek2018distributed} proposed a multi-camera system to classify parking spaces as vacant or occupied. In this solution, the acquired images are processed on-board of Raspberry Pi devices. The extracted information about the status of parking spaces is then transmitted to a central server, which evaluates the parking spaces in the overlapping areas. Their algorithm is based on the histogram of oriented gradients (HOG)\cite{dalal2005histograms} feature descriptor and support vector machine (SVM) classifier. Since the HOG feature descriptor cannot adequately describe rotated vehicles, the authors have provided a descriptor with additional information about rotation to increase the system accuracy.

However, these solutions rely on prior knowledge of the monitored scene, like the position of the parking spaces or some geometric information of the scene. In essence, a preliminary annotation of the new areas and a new training phase of the algorithm are often mandatory operations. As a consequence, these techniques are not very flexible. On the other hand, we propose a simple yet effective solution that does not need any extra information about the monitored scene. The smart cameras can automatically localize and count the vehicles present in their field of view, propagating the single results to the other edge devices through messages. A decentralized technique, again running directly on the edge devices, is instead in charge of analyzing and merging these results, exploiting the captured images' geometry, and automatically outputs the number of cars present in the entire parking area.

\section{Proposed approach}
\label{sec:proposed_solution}

\subsection{Overview}
In this section, we describe our multi-camera counting algorithm. We based our system on the parallel processing of each of the smart cameras followed by the fusion of their results to estimate the number of vehicles present in the \textit{entire} parking area. 

\begin{figure}[htbp]
\centering
  \begin{subfigure}{0.44\textwidth}
    \includegraphics[width=\textwidth]{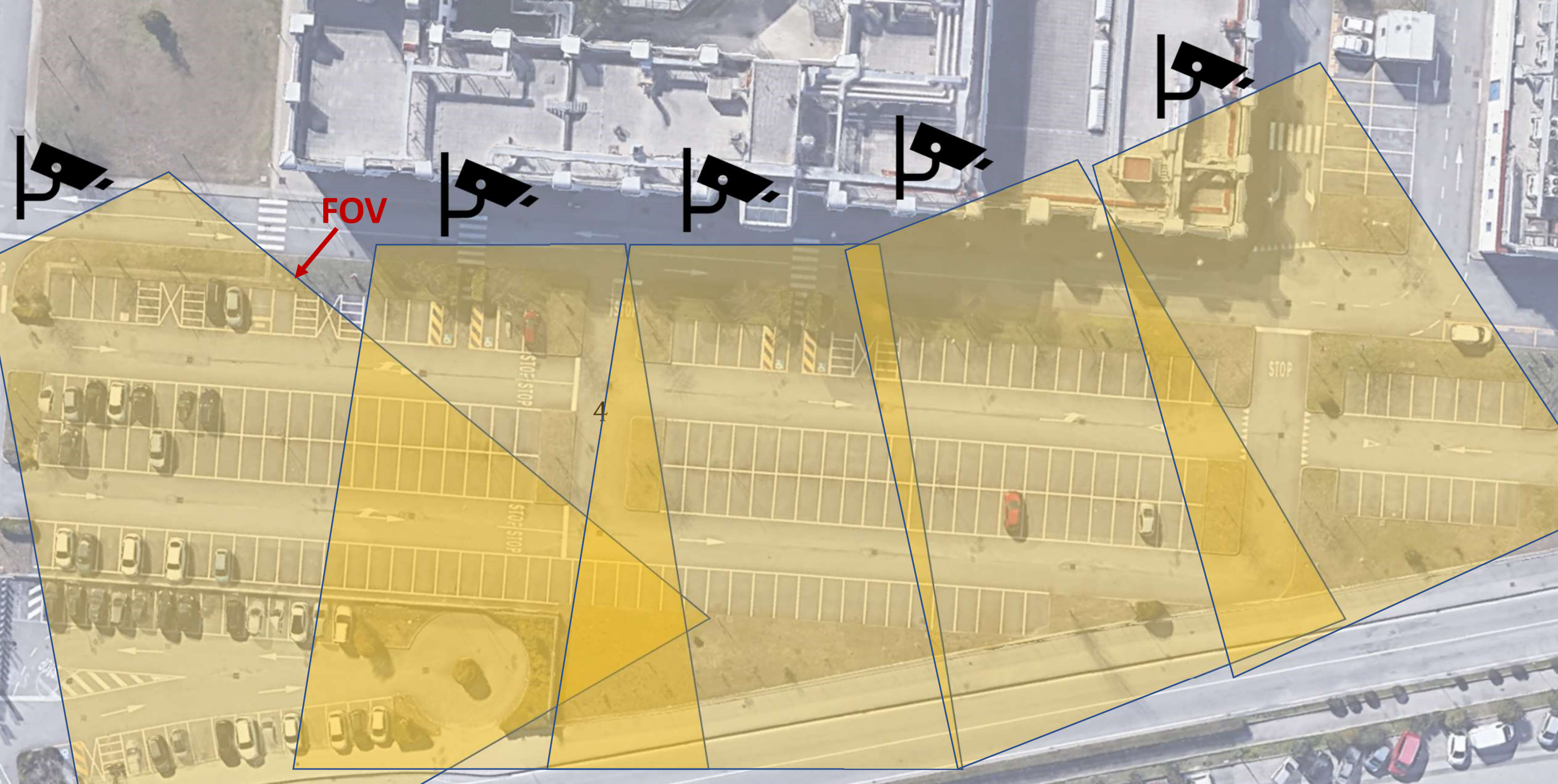}
    \label{system_overview_example}
  \end{subfigure} \hfill
  \begin{subfigure}{0.55\textwidth}
    \includegraphics[width=\textwidth]{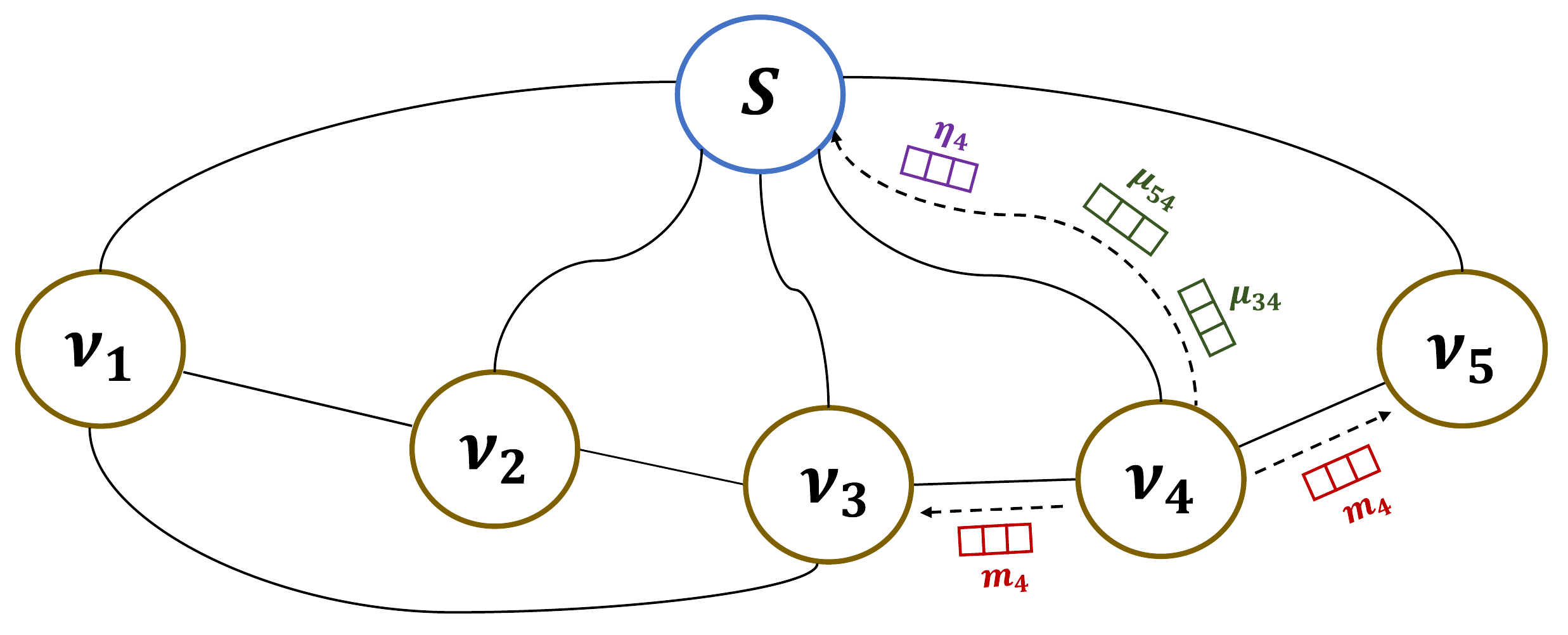}
    \label{system_model_example}
  \end{subfigure}
\caption{An example of our multi-camera counting system, with $n=5$ smart cameras. We model it as a graph $G$, comprised of $n$ nodes $\nu_i$ (one for each camera) and one Sink node $S$, $V = \{\nu_1, \nu_2, …, \nu_n, S\}$. Each node $\nu_i$ can capture images, localize and count the vehicles present in its FOV, and communicate with its neighboring nodes through messages $m_i$ containing these detections. Moreover, each node $\nu_i$ can run a local counting algorithm in charge of computing partial counting results concerning the overlapped areas between its FOV and the ones belonging to its neighbors, exploiting images' geometry. These partial results are sent through messages to the Sink node $S$, which is responsible for their fusion and provides the final result. Messages to $S$ can be of two types: i) $\eta_i$ containing the number of cars captured by the node $\nu_i$ in its FOV, and ii) $\mu_{j, i}$ representing the partial counting estimation related to the overlapping area between two neighboring nodes $\nu_i$ and $\nu_j$.}
\label{fig:system_overview}
\end{figure}

Figure \ref{fig:system_overview} shows an example of our multicamera-camera counting system, together with its graphical representation. We model our system as a graph $G$, comprised of $n$ nodes $\nu_i$ and one Sink node $S$, $V = \{\nu_1, \nu_2, …, \nu_n, S\}$. Each node $\nu_i$ represents an independent edge device, i.e., a smart camera in our case. Two nodes $\nu_i$ and $\nu_j$ are considered neighbors if their FOVs overlap, and in this case, a directed edge of the graph connects them. Each edge device $\nu_i$ can capture images, localize and count the vehicles present in its FOV exploiting a deep learning-based detector, and communicate with its neighboring nodes through messages $m_i$ containing the cars detections. Furthermore, each node $\nu_i$ can also run a local counting algorithm in charge of computing partial counting results concerning the estimation of the number of vehicles present in overlapped areas between its FOV and the ones belonging to its neighbors.

The fusion of the partial results is performed by the Sink node $S$, which is also in charge of providing the final result and synchronizing all the algorithm steps through synchronization signals headed towards the other nodes $\nu_i$. On the other hand, the nodes $\nu_i$ can also communicate through messages with the Sink node. They can be of two types: i) messages $\eta_i$ containing the number of cars captured by the node $\nu_i$ in its FOV, and ii) messages $\mu_{j, i}$ representing the partial counting estimation related to the overlapping area between two neighboring nodes $\nu_i$ and $\nu_j$.

In the following sections, we describe in detail all the steps of our algorithm. First, in Section \ref{sec:system_init}, we outline the automatic system initialization, performed by the smart-cameras themselves, in which they compute a homography between the scene they are monitoring and the scene observed by the neighboring cameras. Then, in Section \ref{sec:local_counting}, we describe the local counting algorithm that runs on each of the smart cameras. It combines a CNN-based counting technique that can localize and estimate the number of vehicles present in their monitored scenes, together with a geometric-based approach responsible for estimating the number of vehicles present in the overlapping areas between the nodes and their neighbors. Finally, in Section \ref{sec:global_counting}, we depict the global counting algorithm responsible for the fusion of these individual and partial results, and that finally outputs the number of cars present in the \textit{entire} parking area.

\subsection{Initialization}
\label{sec:system_init}
This step is aimed at \textit{automatically} initializing the system, estimating the geometric relationship between each node (i.e., each scene monitored by a smart camera) and its neighbors. The only hypotheses we impose are i) each smart camera is aware of the IP addresses of its neighbors, i.e., the cameras having the field of view overlapped with its own; ii) the Sink node $S$ is aware of the IP addresses of all the smart cameras belonging to the system.

The Sink node $S$ starts the initialization phase, sending a synchronization signal to all the other nodes. Once received, each smart camera captures an image of the scene it monitors and sends it to all its neighbors. Once a smart camera $i$ receives an image from a neighboring camera $j$, it computes a homographic transformation $H_{j, i}$ between the image $j$ and the image $i$ describing its monitored scene. This allows us to establish a correspondence between the points belonging to the pair of images taken by the two cameras, which will be used subsequently in the algorithm. We formalized the system initialization for a generic node $\nu_i$ in the Algorithm \ref{alg:system_init}.

\begin{figure}[htbp]
\centerline{\includegraphics[width=.80\textwidth]{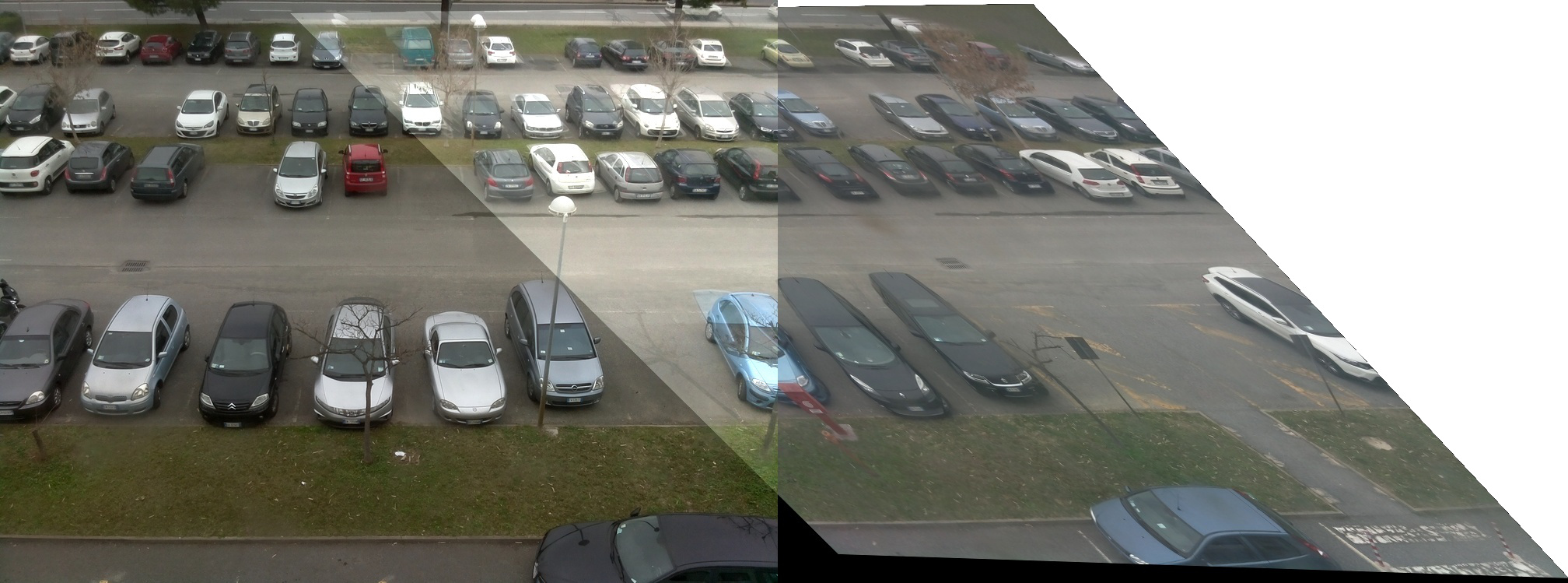}}
\caption{Example of concatenation of two images using a homographic transformation, where it is also visible the overlapping area between them.}
\label{fig:stiching_example}
\end{figure}

However, finding this homography can be challenging because neighboring cameras can have different angles of view, leading to a perspective distortion between the images captured by them. Given a pair of neighboring nodes $\nu_i, \nu_j$, we employ a procedure that starts in finding the SIFT \cite{lowe1999object} key-points and feature descriptors of the images $i, j$ captured by the two nodes. Then, we match the two sets of feature descriptors performing the David Lowe’s ratio test \cite{lowe2004distinctive}, and we further filter the matched feature descriptors by keeping only the pairs whose euclidean distance is below a given threshold. Finally, we apply a random sample consensus (RANSAC \cite{fischler1981random}) to these filtered feature descriptors. Figure \ref{fig:stiching_example} shows the concatenation of two neighboring images $i$ and $j$ in which we apply the found homographic matrix to the image $i$, to have the same perspective as the image $j$. 

\begin{algorithm}[htbp]
\caption{\textbf{: Initialization} \\ At each Initialization Signal by $S$, each node $\nu_i$ performs the following steps:}
\begin{algorithmic}[1]
\State {\Call{ReceiveInitSignal()}{}} \Comment{waits the initialization signal from $S$}
\State {image$_i \gets$ \Call{CameraCapture()}{}}
\ForEach {$j \in J $} \Comment{$J$ is the set of neighboring nodes of node $\nu_i$}
\State {\Call{SendImage}{image$_i$,$\nu_j$}} \Comment{sends image$_i$ to node $\nu_j$}
\State {image$_j \gets$ \Call{ReceiveImage()}{}} \Comment{receives image$_j$ from node $\nu_j$}
\State {$H_{j,i} =$ \Call{ComputeHomography}{image$_j$, image$_i$}}
\EndFor
\end{algorithmic}
\label{alg:system_init}
\end{algorithm}  

\subsection{Local Counting Algorithm}
\label{sec:local_counting}
This section describes the local counting algorithm that runs directly onboard the edge devices. It combines a CNN-based counting technique in charge of the localization and the estimation of the number of vehicles present in the acquired single images, i.e., the contents of the messages $m_{i}$ and the quantities $\eta_i$ shown in Figure \ref{fig:system_overview}, together with a geometric-based approach responsible of estimating the number of vehicles present in the overlapping areas between the nodes and their neighbors, i.e., the quantities $\mu_{j, i}$ in Figure \ref{fig:system_overview}.

\paragraph{A vehicle counting CNN on the Edge}
We base our vehicle counting technique on \textit{Mask R-CNN} \cite{he2017mask}, a popular deep CNN for instance segmentation that operates within the `recognition using regions' paradigm \cite{gu2009recognition}. In particular, it extends the \textit{Faster R-CNN} detector \cite{Ren2015} by adding a branch that outputs a binary mask saying whether or not a given pixel is part of an object. Briefly, in the first stage, a CNN acts as a backbone, extracting the input image features. Starting from this feature space, another CNN named Region Proposal Network (RPN) generates region proposals that might contain objects. RPN slices pre-defined region boxes (called anchors) over this space and ranks them, suggesting those most likely containing objects. Once RPN produces the Regions Of Interests (ROIs),  they might be of different sizes. Since it is hard to work on features having different sizes, RPN reduces them into the same dimension using the Region of Interest Pooling algorithm. Finally, these fixed-size proposals are processed by two parallel CNN-based branches: one is responsible for classifying and localizing the objects inside them with bounding boxes; the second outputs a binary mask that says whether or not a given pixel is part of an object. In the end, given an input image, the network produces per-pixel masks localizing the detected objects together with the associated labels classifying them. 

To make our counting solution able to run efficiently directly on the edge devices, we employ, as a backbone, the \textit{ResNet50} architecture, a lighter version of the popular \textit{ResNet101} \cite{he2016deep}. This simplification is also justified because the more powerful version of Mask R-CNN based on the ResNet101 model was designed for more complicated visual detection tasks than ours. Originally, Mask R-CNN was trained on the \textit{COCO} dataset \cite{lin2014microsoft} to detect and recognize 80 different classes of everyday objects. In our case, we have to localize and identify objects belonging to just one category (i.e., the \textit{vehicle} category). To this end, we further simplify the model by reducing the number of the final fully convolutional layers responsible for the classification of the detected objects, making the model lighter. Once we have localized the instances of the objects, we count them estimating the number of vehicles present in the scene.

\paragraph{Local counting} The Sink node $S$ starts this phase, sending a synchronization signal to all the smart cameras belonging to the system. Once received the synchronization signal, each node $\nu_i$ captures an image belonging to its underlying FOV, feeding the previously described CNN-based counting technique with it and obtaining as output a set of masks masks$_i$ localizing the vehicles present in the scene. The cardinality of this set of masks corresponds to the number of cars present in the image, i.e., the quantity $\eta_i$, that is sent through a message to the Sink node $S$. Then, the node $\nu_i$ packs this set of masks masks$_i$ in a message $m_i$, sending it to all its neighboring nodes $\nu_j$, and receiving from them their corresponding set of masks masks$_j$ packed in a message $m_j$. Once received a message $m_j$, the node $\nu_i$ is responsible for analyzing the potential vehicles present in the overlapped area between its FOV and the one of the node $\nu_j$. To this end, it employs the homographic transformation $H_{j, i}$ computed during the system initialization, as described in Section \ref{sec:system_init}, projecting the masks belonging to the set masks$_j$ into its image plane, filtering them and discarding the ones that overlap with the masks belonging to the set masks$_i$ having a value of Intersection over Union (IoU) greater than a threshold that we empirically found to be optimal at 0.2. These masks indeed localize vehicles already detected, and that should not be considered a second time. On the other hand, the cars left after this filtering are vehicles that were not detected in the FOV underlying the node $\nu_i$, but instead found by the node $\nu_j$, probably because of having a better view of this object. Referring to our graph modeling the system and reported in Figure \ref{fig:system_overview}, the number of the discarded cars after this filtering operation corresponds to the message $\mu_{j, i}$, that is sent to the Sink node $S$. We detail all the described steps in the Algorithm \ref{alg:local_counting} and in the Procedure \ref{alg:compute_num_overlaps}.

\begin{algorithm}[htbp]
\caption{\textbf{: Local Counting} \\ At each Computational Signal by $S$, each node $\nu_i$ performs the following steps:}

\begin{algorithmic}[1]
\State {\Call{ReceiveComputSignal()}{}} \Comment{waits the computational signal from $S$}
\State {image$_i \gets$ \Call{CameraCapture()}{}}

\State {masks$_i \gets$ \Call{MaskRCNN}{image$_i$}}
\State {$\eta_i \gets \left| \text{masks}_i\right |$}
\State {\Call{SendMessage}{$\eta_i, S$}} \Comment{sends $\eta_i$ to Sink node $S$}
\State {$m_i \gets$ \Call{PackMessage}{masks$_i$}} \Comment{builds message $m_i$ containing masks$_i$}
\ForEach {$j \in J $} \Comment{$J$ is the set of neighboring nodes of node $\nu_i$}
\State {\Call{SendMessage}{$m_i, \nu_j$}} \Comment{sends $m_i$ to node $\nu_j$}
\State {$m_j \gets$ \Call{ReceiveMessage()}{}} \Comment{receives message $m_j$ from node $\nu_j$}
 \State {masks$_j \gets$ \Call{UnpackMessage}{$m_j$}} \Comment{unpacks $m_j$ containing masks$_j$}
\State {$\mu_{j, i} \gets$ \Call{compute\_$\mu$}{masks$_i$, masks$_j$, $H_{j, i}$}}
\State {\Call{SendMessage}{$\mu_{j, i}, S$}} \Comment{sends $\mu_{j, i}$ to Sink node $S$}
\EndFor
\end{algorithmic}

\label{alg:local_counting}
\end{algorithm}  

\begin{algorithm}[htbp]
\caption{: \textbf{Computation of $\mu$} \\ $\mu$ represents the num of cars detected by $\nu_j$ and already detected by $\nu_i$ \\ Each node $\nu_i$ performs the following procedure:}

\begin{algorithmic}[1]
\Procedure{compute\_$\mu$}{masks$_i$, masks$_j$, $H_{j,i}$}

\State {n\_cars\_already\_detected $\gets 0$}
\ForEach {mask $\in$ masks$_j$}
\State  {mask$_h \gets $} \Call{Project}{$H_{j,i}$, mask} \Comment{projects mask points on plane $i$}
\If {mask$_h$ falls within image$_i$}
\State{mask$_\text{max} \gets \argmax_{m \in \text{masks}_i} \text{IoU}(\text{mask}_h, m)$}
 \If {IoU$(\text{mask}_h, \text{mask}_\text{max}) > \tau$ }
 \State {n\_cars\_already\_detected ++}
 \EndIf
\EndIf
\EndFor
\State {\textbf{return} n\_cars\_already\_detected}
\EndProcedure
\end{algorithmic}
\label{alg:compute_num_overlaps}
\end{algorithm}

\subsection{Global Counting Algorithm}
\label{sec:global_counting}
In this section, we describe the global counting algorithm that runs on the Sink node $S$, responsible for the fusion of the partial results coming from all the other nodes and that finally outputs the number of cars present in the \textit{entire} monitored parking area.

This phase starts when $S$ receives all the $\eta_i$ and the $\mu_{j, i}$, i.e., the number of vehicles estimated in the single FOVs and the estimation of the number of cars already considered in the overlapping areas between neighboring cameras, from all the nodes belonging to the system. In particular, for each overlapped area shared between a pair of nodes $\nu_i, \nu_j$, the node $S$ receives two messages $\mu_{j, i}$ and $\mu_{i, j}$, the contents of which are computed by the two nodes employing two homographic transformations $H_{j, i}$ and $H_{i, j}$, respectively. These two quantities can be potentially different. We choose the best value aggregating them, choosing between three different functions - max, min and mean, finding that the latter is the best one. Finally, the node $S$ builds the final result, i.e., the estimation of the number of vehicles present in the \textit{entire} parking lot, by summing up all the $\eta_i$, and subtracting the aggregated values. We detail all these steps in the Algorithm \ref{algo:global_counting}.

\begin{algorithm}[htbp]
\caption{\textbf{: Global Counting} \newline The Sink node $S$ performs the following steps:}

\begin{algorithmic}[1]
\ForEach {$(\mu_{i, j}, \mu_{j, i})$}
\State{$\overline{\mu_k} \gets$ \Call{Aggregate}{$\mu_{i, j}, \mu_{j, i}$}}
\EndFor
\State{global\_cars\_count $\gets \sum_{n=1}^{N} \eta_n - \sum_{k=1}^{K} \overline{\mu_k}$ \newline \Comment{$N$ is the set of nodes, $K$ is the set of aggregations}}
\end{algorithmic}

\label{algo:global_counting}
\end{algorithm}

\section{Experimental Setup}
\label{sec:exp_setup}

\subsection{The CNRPark-EXT Dataset}
\label{sec:datasets}

In this work, we exploit the \textit{CNRPark-EXT} public dataset introduced in \cite{amato2017deep}, a collection of annotated images of vacant and occupied parking spaces in the campus of the National Research Council (CNR) in Pisa, Italy. This dataset is challenging and describes most of the problematic situations that can be found in a real scenario: nine different cameras capture the images under various weather conditions, angles of view, light conditions, and many occlusions. Furthermore, the cameras have their fields of view partially overlapped. Since this dataset is specifically designed for parking lot occupancy detection, it is not directly usable for the counting task. Indeed, each image, called \textit{patch}, contains one parking space labeled according to its occupancy status - 0 for vacant and 1 for occupied. Since this work aims at counting the cars present in the parking area, we extended it by considering the full images and adapting the ground truth to our purposes. 

To train and evaluate the vehicles counting CNN based on Mask R-CNN, we created a suitable label set. In this case, these labels correspond to \textit{binary masks}, i.e., binary images identifying the polygons surrounding the vehicles we want to detect. Since mask creation is a very time-consuming operation, differently from our previous work \cite{ciampi2018counting}, we considered the \textit{raw} masks obtained directly from the bounding boxes localizing the occupied parking spaces. The idea is that we do not need precise polygons that identify the vehicles we want to detect. Still, we can use the region within the delimiters that identify the occupied parking spaces and the underlying part of the car.

On the other hand, to validate our multi-camera algorithm, we considered some sequences of images belonging to different cameras captured simultaneously. In other words, we took into account some snapshots of the whole parking area picked up by the different views of the multiple cameras. We manually annotated these sequences, counting the vehicles present in the scenes, considering them just once, and discarding them from the global count if they were located in the overlapping areas. In particular, we accounted for six different sequences, two for each weather condition, considering the images belonging from camera$_2$ to camera$_9$. We did not consider camera$_1$ since it has a very different view of the parking area compared to the other ones, covering a big portion of the whole parking lot already monitored by the remaining cameras, and employing a very large perspective resulting in capturing images with very small cars that sometimes are not well distinguishable. 

\subsection{Evaluation Metrics}
\label{sec:metrics}
Following other counting benchmarks, we exploit Mean Absolute Error (\textit{MAE}), Mean Square Error (\textit{MSE}), and Mean Relative Error (\textit{MRE}) as the metrics for the performance evaluation, defined as follows:
\begin{equation}
MAE = \frac{1}{N} \sum_{n=1}^{N} |c_n^{gt} - c_n^{pred}|
\end{equation}

\begin{equation}
MSE = \frac{1}{N} \sum_{n=1}^{N} (c_n^{gt} - c_n^{pred})^{2}
\end{equation}

\begin{equation}
MRE = \frac{1}{N} \sum_{n=1}^{N} \frac{|c_n^{gt} - c_n^{pred}|}{ \textrm{num\_spaces}_n}
\end{equation}

where $N$ is the total number of the images, $c_{gt}$, $c_{pred}$ and $num\_spaces_n$ are the actual count, the predicted count, and the total number of parking spaces of the n-th image, respectively. Note that as a result of the squaring of each difference, MSE effectively penalizes large errors more heavily than small ones. Then MSE should be more useful when large errors are particularly undesirable. On the other hand, MRE also considers the relation between the error and the total number of objects present in the image.

\subsection{Implementation Details}
We report in this section some implementation details concerning the Mask R-CNN-based algorithm responsible for the prediction of the number of vehicles in the single images. In particular, we trained the modified Mask R-CNN initializing the weights of the ResNet50 backbone with the ones of a pre-trained model on \textit{ImageNet} \cite{deng2009imagenet}, a popular dataset for classification tasks, and the remaining ones at random. We freeze the backbone for the firsts 10 epochs, and then we trained the whole network for 20 additional epochs. To prevent overfitting, we applied some standard augmentation techniques to the training data: images are horizontally flipped with a 0.5 probability, then their pixels are multiplied by a random value between 0.8 and 1.5, and finally, they are blurred using a Gaussian kernel with a standard deviation of a random value between 0 and 5. Then, to support training multiple images per batch, we resized all pictures to the same size. If an image is not square, we pad it with zeros to preserve the aspect ratio. In the end, we obtained images of size $1024\times1024$. At inference time, images are resized and padded with zeros to get a square picture of size $1024\times1024$, and no other augmentations take place.

\section{Experiments and Results}
\label{sec:experiments}
In this section, we report the experiments and the obtained results. Firstly, we evaluate the performance against other state-of-the-art solutions of the CNN-based technique responsible for estimating the vehicles in the single images directly onboard the smart cameras, also stressing its generalization capabilities. Then, we validate the effectiveness of our multi-camera algorithm, demonstrating that our system can benefit from the redundant information deriving from the different cameras.

\subsection{Experiments on the CNN-based counting solution on the edge}

\subsubsection{State-of-the-art comparison}

\begin{figure}[htbp]
\centering
  \begin{subfigure}[b]{0.48\textwidth}
    \includegraphics[width=\textwidth]{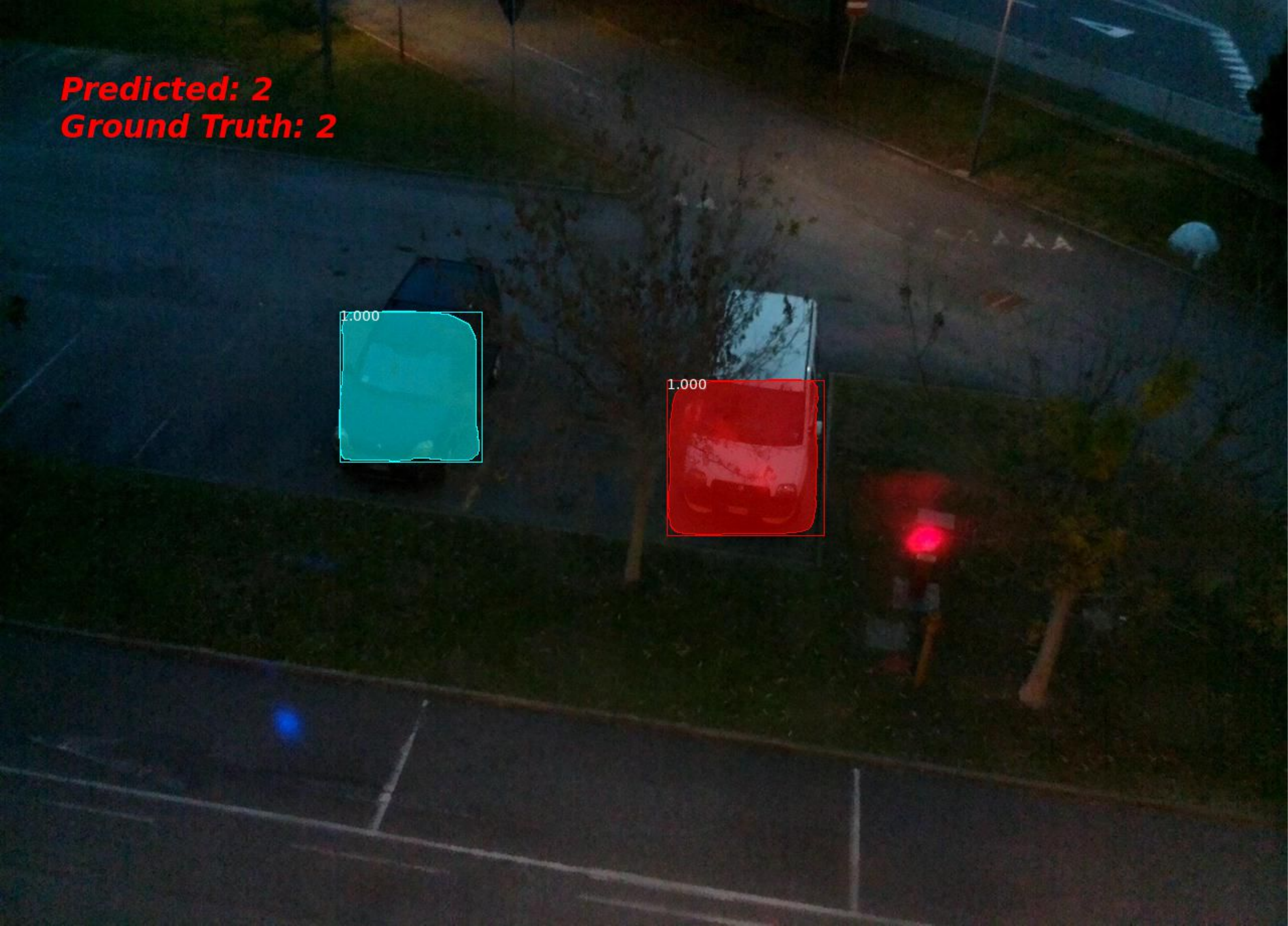}
    \caption{Image from Camera$_2$}
    \label{example_counting_output_a}
  \end{subfigure} \hfill
  \begin{subfigure}[b]{0.48\textwidth}
    \includegraphics[width=\textwidth]{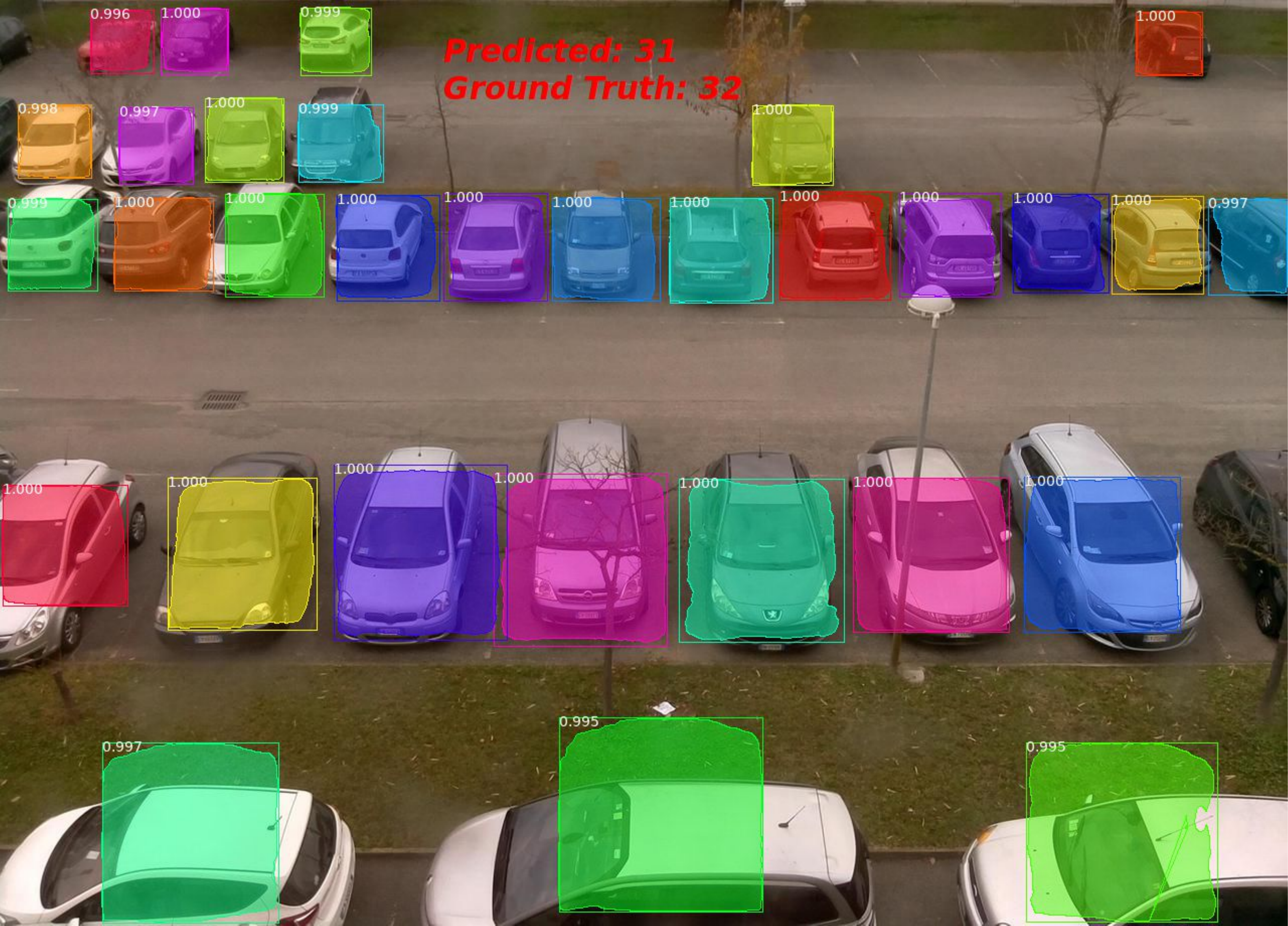}
    \caption{Image from Camera$_8$}
    \label{example_counting_output_b}
  \end{subfigure}
  \caption{Two examples of the output of our counting method. Images are taken from the CNRPark-EXT dataset. We report the predictions and the estimate of the number of vehicles present in the scene.}
  \label{fig:example_counting_output}
\end{figure}

We compare our solution with the results obtained in our previous work \cite{ciampi2018counting}, where we presented a centralized counting approach based on the original version of Mask R-CNN having the ResNet101 model as features extractor, which has been fine-tuned on a very small manually annotated subset of the CNRPark-EXT dataset, starting from the model pre-trained on the \textit{COCO} \cite{lin2014microsoft} dataset. We filter the detections considering only the predictions related to the car class, and we count them. Although this solution is very computationally expensive and unsuitable for edge devices, it represents a direct comparison in terms of counting on the same dataset. We also compare our technique against the method proposed in \cite{amato2017deep}, an approach for car parking occupancy detection based on \textit{mAlexNet}, a deep CNN designed explicitly for smart cameras. This work represents an indirect method for counting cars in a car park, as the counting problem is cast as a classification problem: if a parking space is occupied, we increment the total number of cars, otherwise not. We illustrate the results in Table \ref{tab:results_edge_counting}, where we also report the performance obtained using the Mask R-CNN network without a preliminary fine-tuning on the CNRPark-EXT dataset. Our solution performs better than the other state-of-the-art considered methods, considering all three counting metrics. In particular, our approach outperforms the solution introduced in \cite{ciampi2018counting}, despite the latter employs a more deep and powerful CNN, and it is designed to be used as a centralized-server solution. This is explained by the fact that in \cite{ciampi2018counting} the authors fine-tuned the CNN using a tiny dataset. Consequently, the algorithm overfits on the training data, and it cannot generalize over the test subset. It is also worthy of notice that our CNN also outperforms the mAlexNet network, even though the latter knows the exact location of the parking spaces. Figure \ref{fig:example_counting_output} shows some examples of images belonging to different cameras and different weather conditions together with the masks localizing them computed by our counting solution. 

\begin{table}[htbp]
\caption{Results obtained using our counting solution on the edge, compared with other state-of-the-art approaches. We get the best results on all the three considered counting metrics.}
\centering
\begin{tabularx}{\linewidth}{Xrrr}
\toprule
Method & MAE & MSE & MRE\\
\midrule
mAlexNet \cite{amato2017deep}                              & 1.34 & 8.00 & 0.04\\
Fine-Tuned ResNet101 Mask R-CNN \cite{ciampi2018counting}  & 1.05 & 4.41 & 0.03\\
ResNet50 Mask R-CNN                                   & 11.20 & 247.40 & 0.30\\
Our solution                                                 & \textbf{0.49} & \textbf{1.04} & \textbf{0.01}\\
\bottomrule
\end{tabularx}
\label{tab:results_edge_counting}
\end{table}

\subsubsection{Generalization capabilities}
Errors in vehicle detection and counting are due to many reasons, but critical points are different light conditions and diverse perspectives. Weather conditions might produce significant illumination changes since puddles and wet floors create a textural pattern that may lead to an error, and sunbeams can create reflections on the car windscreen, covering the majority of the images with saturated patterns. When a CNN does not generalize well, it works well only in the conditions where it was trained. 

To measure the robustness of our approach to these scenarios, we performed two types of experiments: \textit{inter-weather} and \textit{inter-camera} experiments. In the former, we trained our CNN with images taken in one particular weather condition, and we computed the performance metrics obtained on images having different weather conditions. In particular, we performed three experiments, training respectively on the \textit{Sunny}, \textit{Overcast} and \textit{Rainy} subsets of the CNRPark-EXT dataset. In the latter, we trained our algorithm employing images from one camera, and then we computed the performance metrics on pictures captured by another camera. In particular, we performed two experiments, training with images coming respectively from camera$_1$ and camera$_8$. We chose these two cameras because they are particularly representative since one has a side view of the parking lot while the other has a pure front view.

\begin{figure}[htbp]
  \begin{subfigure}[b]{0.31\textwidth}
    \includegraphics[width=\textwidth]{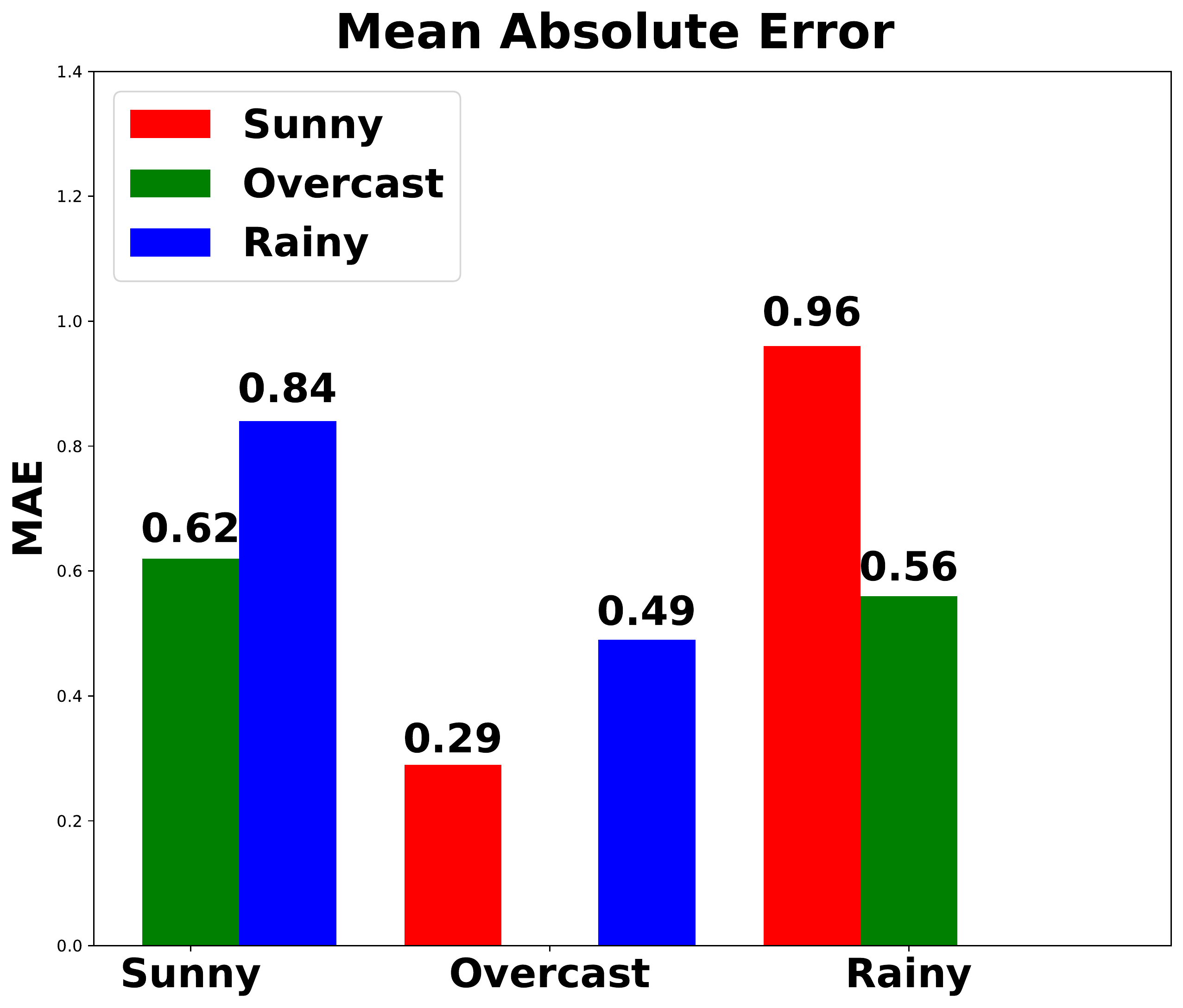}
    \caption{Mean Absolute Error}
    \label{inter_weather_results_a}
  \end{subfigure}
  \hfill
  \begin{subfigure}[b]{0.31\textwidth}
    \includegraphics[width=\textwidth]{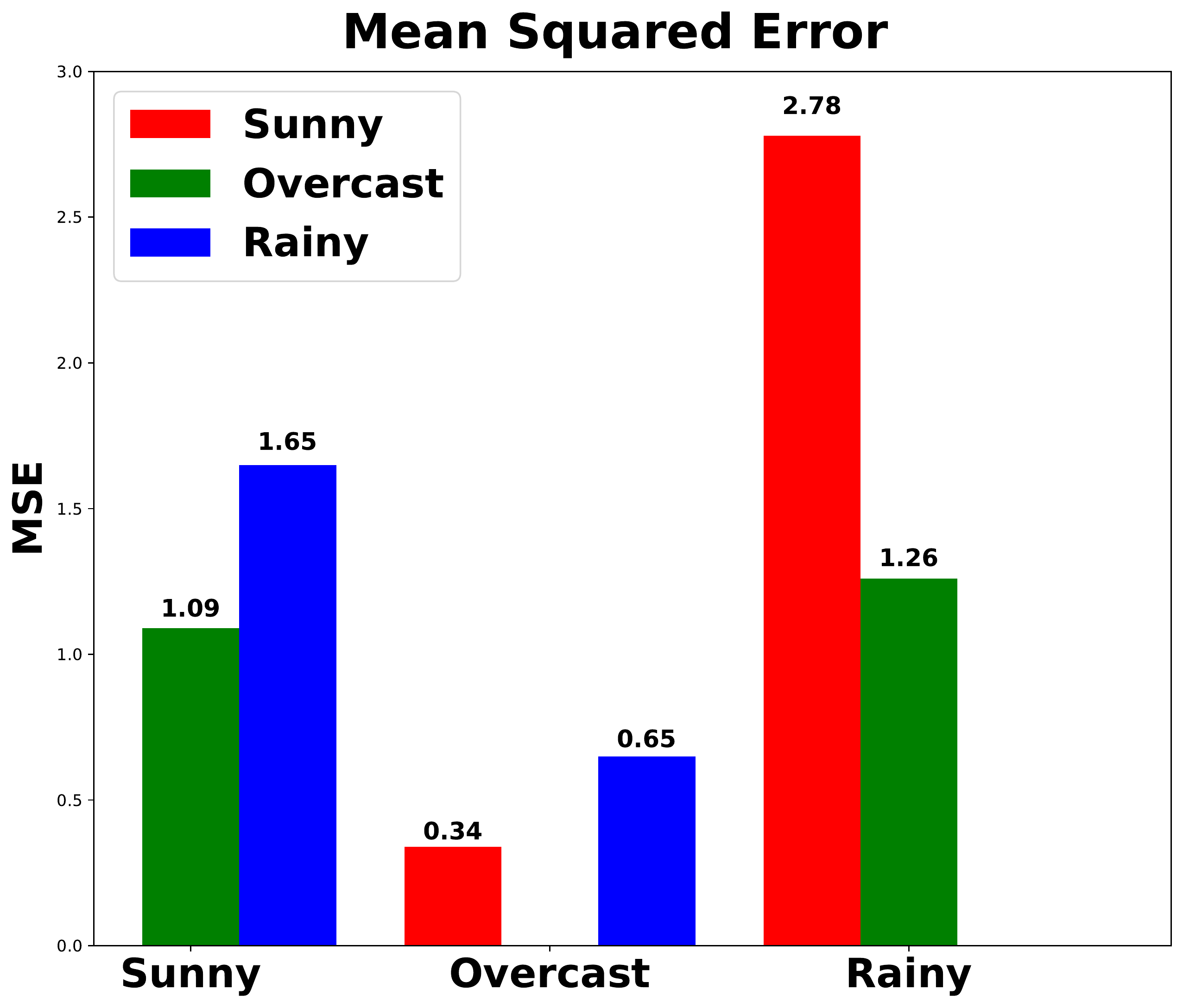}
    \caption{Mean Squared Error}
    \label{inter_weather_results_b}
  \end{subfigure}
  \hfill
  \begin{subfigure}[b]{0.31\textwidth}
    \includegraphics[width=\textwidth]{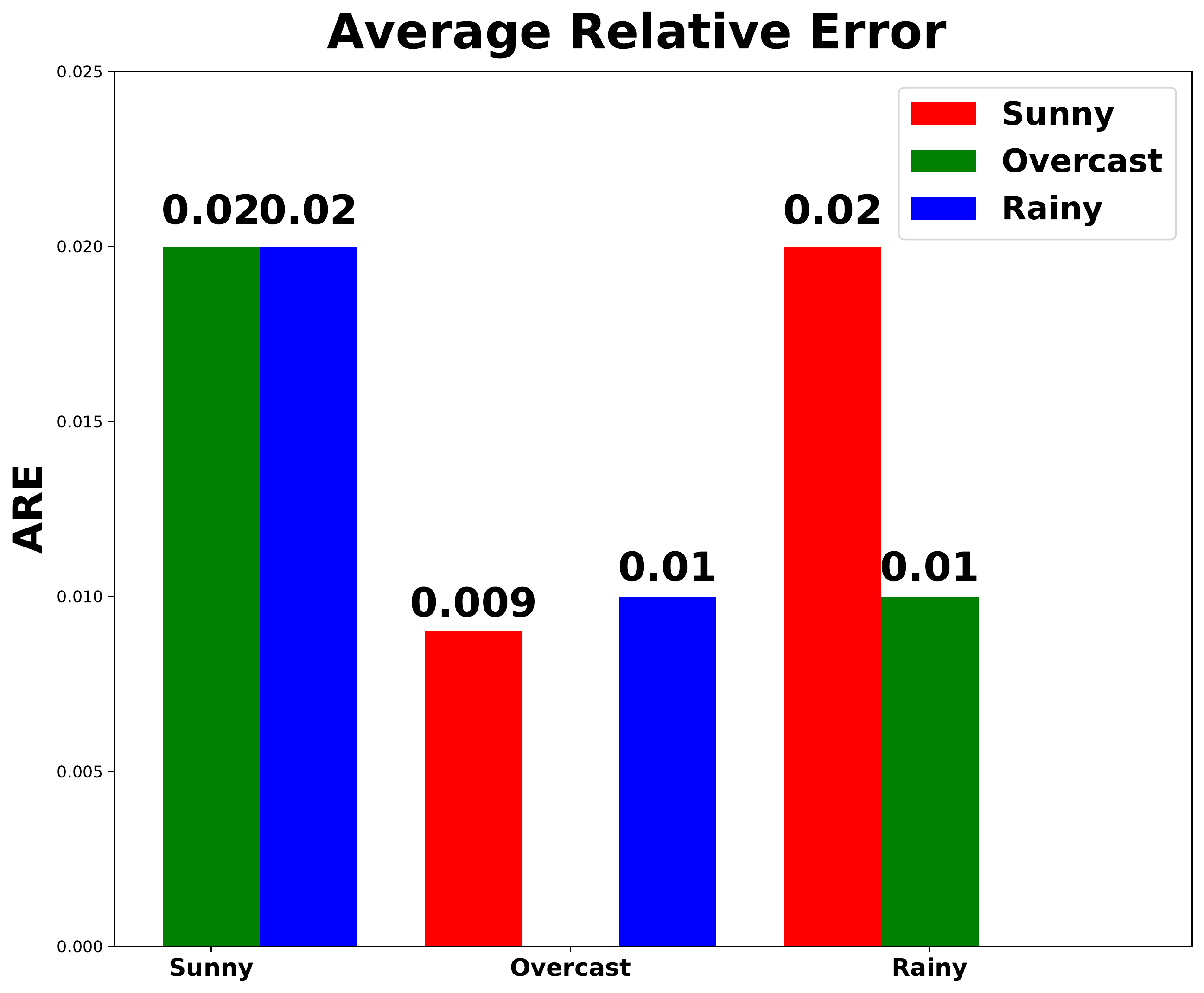}
    \caption{Mean Relative Error}
    \label{inter_weather_results_c}
  \end{subfigure}
  \caption{Results in terms of counting of the inter-weather experiments. The red bar represents the training on sunny images, the blue bar the training on rainy images, and the green one the training on overcast images.}
  \label{inter_weather_results}
\end{figure}

\begin{figure}[htbp]
  \begin{subfigure}[b]{0.31\textwidth}
    \includegraphics[width=\textwidth]{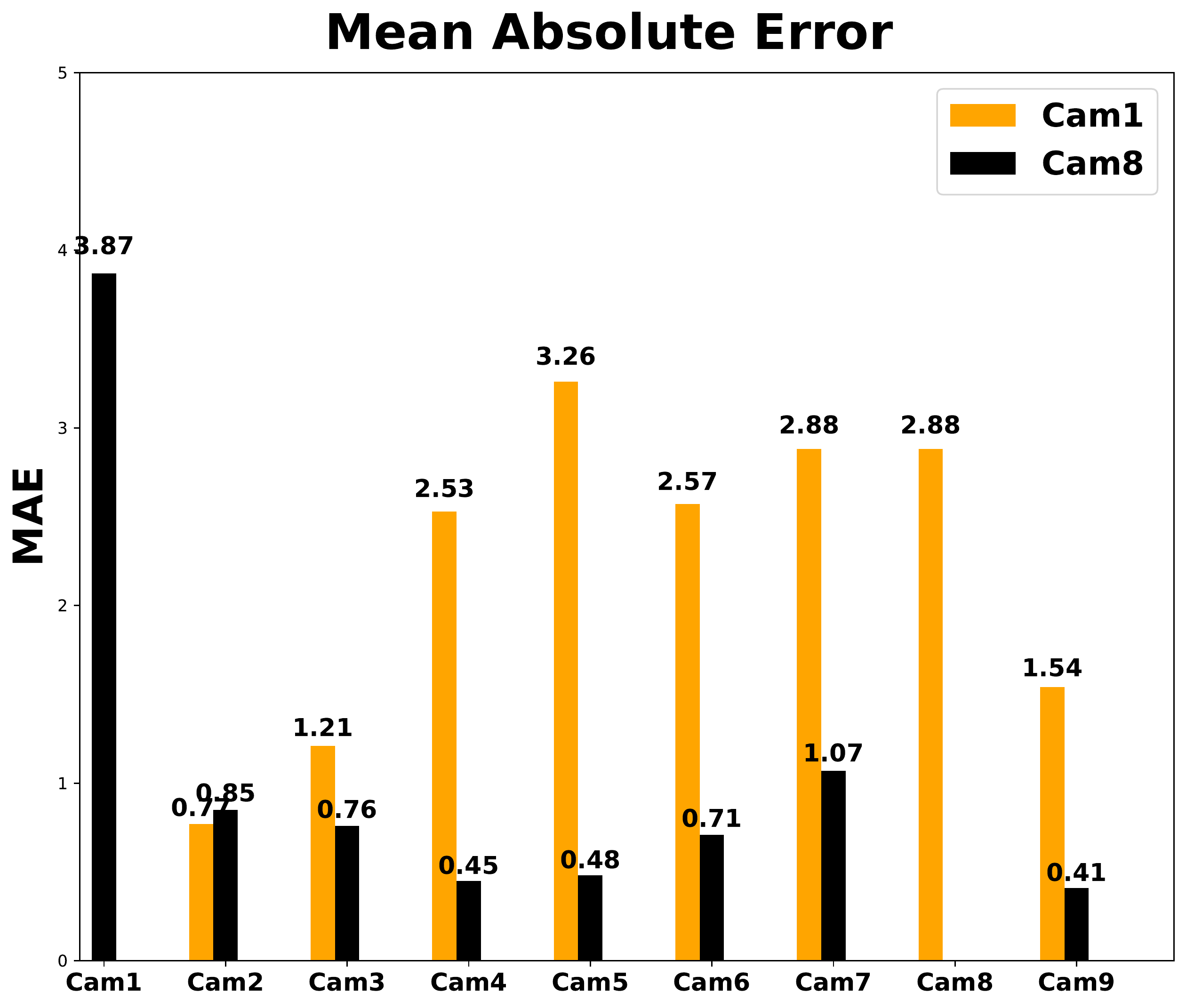}
    \caption{Mean Absolute Error}
    \label{inter_camera_results_a}
  \end{subfigure}
  \hfill
  \begin{subfigure}[b]{0.31\textwidth}
    \includegraphics[width=\textwidth]{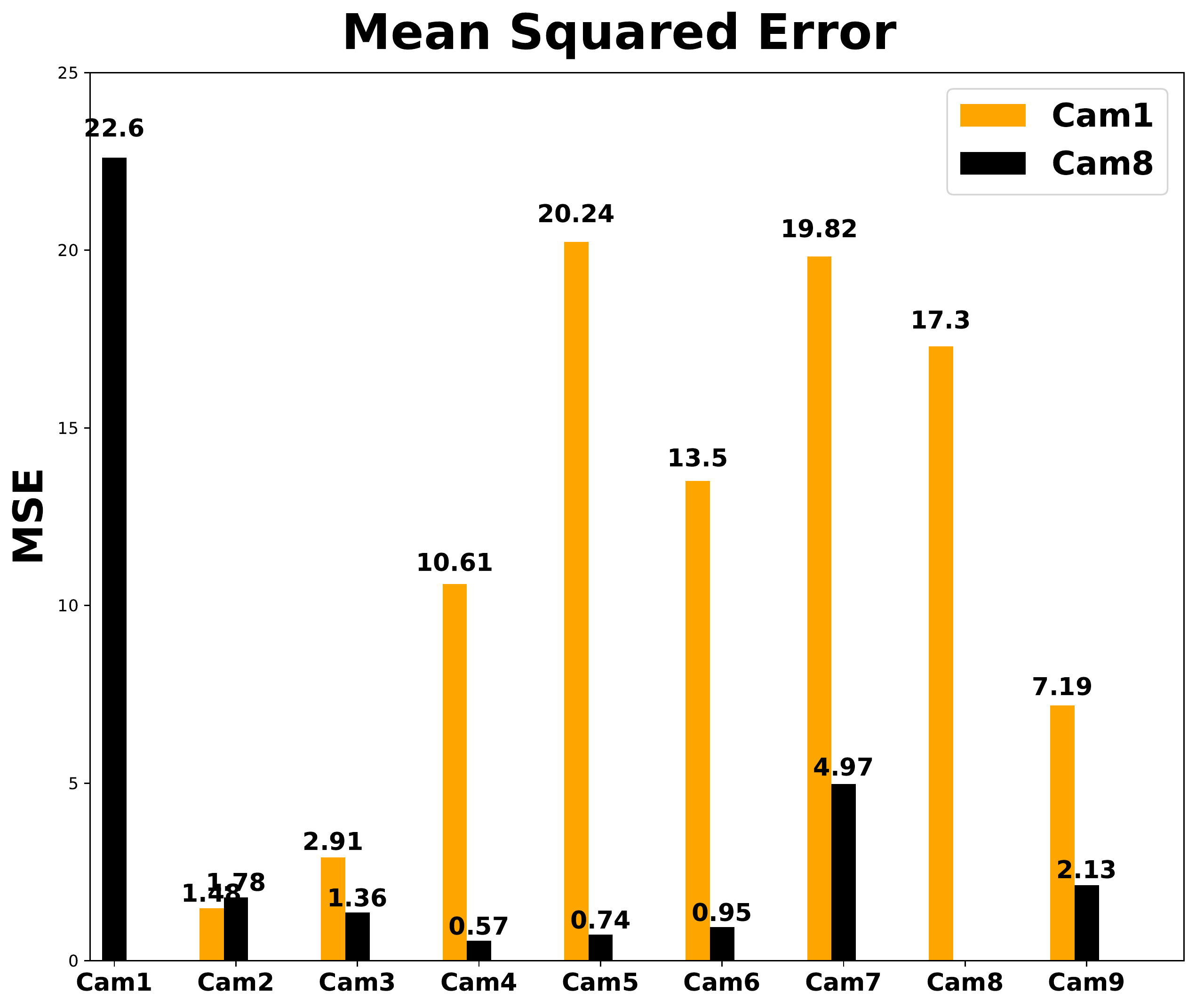}
    \caption{Mean Squared Error}
    \label{inter_camera_results_b}
  \end{subfigure}
  \hfill
  \begin{subfigure}[b]{0.31\textwidth}
    \includegraphics[width=\textwidth]{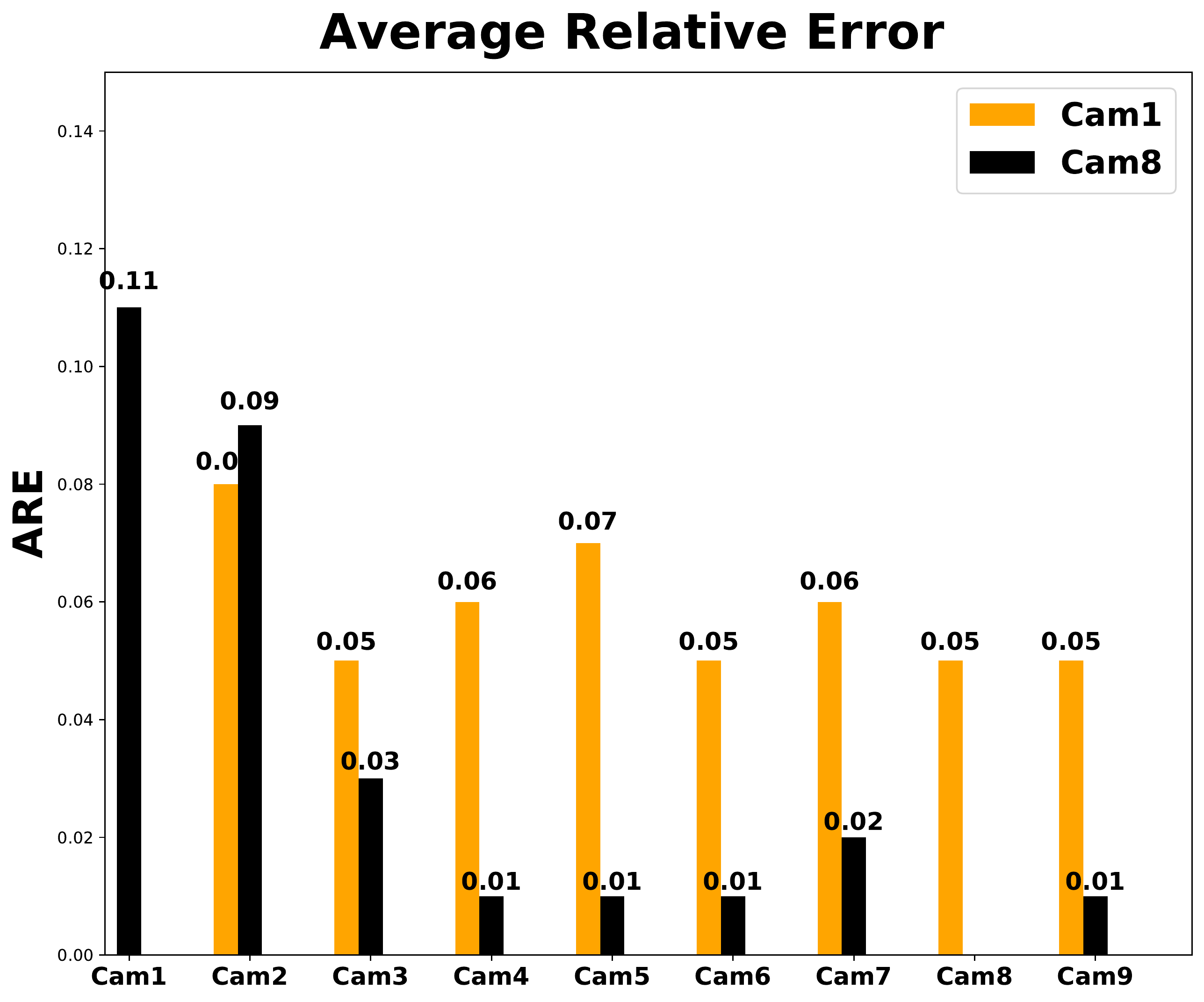}
    \caption{Mean Relative Error}
    \label{inter_camera_results_c}
  \end{subfigure}
  \caption{Results in terms of counting of the inter-camera experiments. The yellow bar represents the training on the Camera$_1$ images while the black bar the training on the Camera$_8$ images.}
  \label{inter_camera_results}
\end{figure}

We report the results of the two experiments in Figure \ref{inter_weather_results} and Figure \ref{inter_camera_results}, respectively. The histograms compare the counting performance metrics of the CNN trained on a specific scenario when tested over all the other possible scenarios. We achieve a good generalization in both the considered scenarios. We experienced a larger amount of error when the CNN is trained and tested on two opposite weather conditions, for instance, \textit{Sunny} and \textit{Rainy}, while the more accurate model was the one trained on \textit{Overcast} weather conditions. However, the performance difference is quite small. On the other hand, in \textit{inter-camera} experiments, the model trained on camera$_8$ is the best, and it has a slight drop in performance only when tested on the camera$_1$ subset. The model trained on the camera$_1$ dataset performs in general worse. This is probably due to a bias in the CNRPark-EXT dataset, where the majority of the images are captured from a frontal viewpoint.

\subsection{Experiments on the Multi-Camera Algorithm}
Due to the absence of other works that consider the CNRPark-EXT dataset as a whole parking area, it is not possible to make a quantitative comparison of our proposed multi-camera algorithm with respect to other state-of-the-art approaches. However, we compare our solution against a baseline and a simplified version of our algorithm, highlighting the improvements obtained by the use of the redundant information deriving from the multiple angles of view.

In particular, we compare our solution against a system that is not aware of the other cameras' overlapped areas, and so it just sums all the vehicles detected by all the cameras belonging to a sequence (Baseline \textbf{B}). Then, we consider a more conservative approach, where the nodes employ the homographic transformations only with the purpose of black-masking the overlapped areas (Simplified algorithm \textbf{S}). This latter baseline then loses the ability to take advantage of monitoring the same lots from different views. However, it is still aware of the locations of the overlapping areas, and it considers the vehicles inside them only once. 

\begin{figure}[htbp]
\centering
  \begin{subfigure}[b]{0.48\textwidth}
    \includegraphics[width=\textwidth]{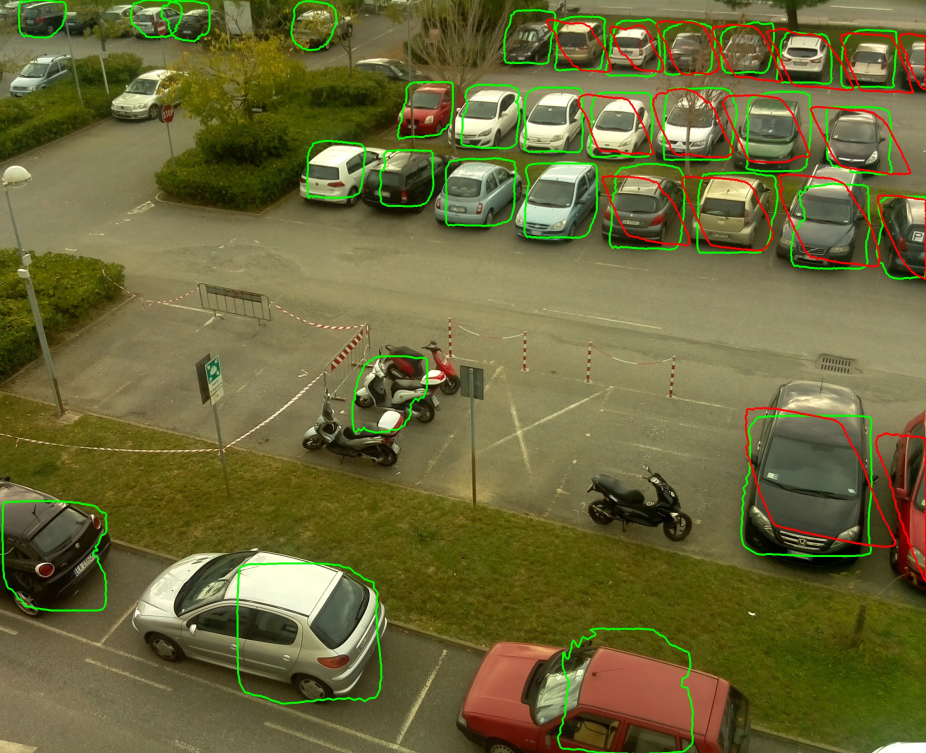}
    \caption{Image from Camera$_9$}
    \label{example_multi_camera_counting_output_a}
  \end{subfigure} \hfill
  \begin{subfigure}[b]{0.48\textwidth}
    \includegraphics[width=\textwidth]{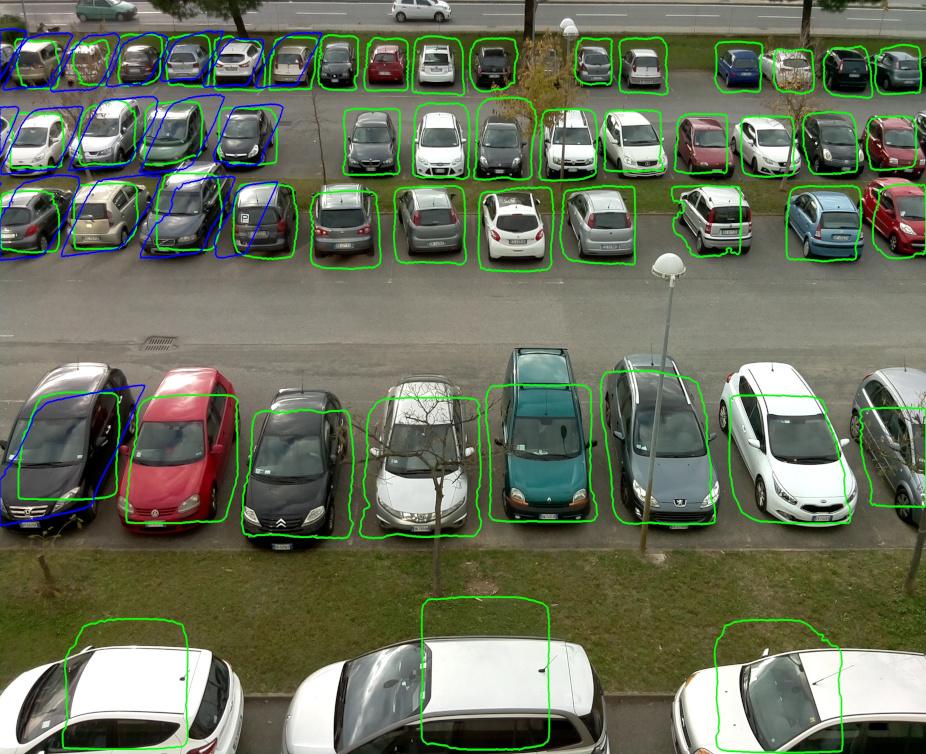}
    \caption{Image from Camera$_8$}
    \label{example_multi_camera_counting_output_b}
  \end{subfigure}
  \caption{Example of the output of our multi-camera algorithm for a pair of images belonging to two different cameras $i, j$ having a shared area in their FOV. We report in green the masks localizing the vehicles detected by a camera in its own FOV, while in red and blue, the masks projected from camera j to camera i and vice-versa, employing the homographic transformations pre-computed during the system initialization.}
  \label{fig:example_multi_camera_counting_output}
\end{figure}

Results are shown in Table \ref{tab:results_multi_camera_counting}. Our solution obtains the best results compared to the considered baselines in all the three counting metrics and all the employed scenarios. We report the errors concerning the considered six sequences of the CNRPark-EXT dataset, together with the MAE, MSE, and MRE, which summarize the mean results regarding all the scenarios. As an example, in Figure \ref{fig:example_multi_camera_counting_output} we also report the output of our multi-camera algorithm for a pair of images belonging to two different cameras having a shared area in their field of view, where we highlight in red and blue the masks projected from one camera to the other, using the previously computed homographic transformations.

\begin{table}[htbp]
\caption{Results using our multi-camera counting algorithm, considering the \textit{entire} parking lot. We compare our solution against a baseline and a simplified version of our algorithm. We report the errors obtained on the six considered sequences (two for each weather condition) of the CNRPark-EXT dataset that we extend on purpose.}
\centering

\footnotesize
\setlength\tabcolsep{.25em}
\begin{tabularx}{\linewidth}{X*{3}{c}@{\hspace{10pt}}*{3}{c}@{\hspace{10pt}}*{3}{c}@{\hspace{10pt}}*{3}{c}}
\toprule
 & \multicolumn{3}{c}{Error} & \multicolumn{3}{c}{Absolute Err.} & \multicolumn{3}{c}{Squared Err.} & \multicolumn{3}{c}{Relative Err. (\%)} \\
 \cmidrule(r){2-4} \cmidrule(r){5-7} \cmidrule(r){8-10} \cmidrule{11-13}
 & B & S & O & B & S & O & B & S & O & B & S & O \\
\midrule
Overcast-1 &        124 &        -33 &    \textbf{2} &      124   &       33   &  \textbf{2}   &    15,376   &     1,089   &   \textbf{4}   &      71.6 &       19.0 &  \textbf{1.2} \\
Overcast-2 &        131 &        -26 &    \textbf{1} &      131   &       26   &  \textbf{1}   &    17,161   &       676   &   \textbf{1}   &      76.1 &      15.1 &  \textbf{0.6} \\
Rainy-1    &         80 &        -39 &   \textbf{-5} &       80   &       39   &  \textbf{5}   &     6,400   &     1,521   &  \textbf{25}   &      47.6 &      23.2 &  \textbf{2.9} \\
Rainy-2    &        105 &        -44 &    \textbf{-5} &      105   &       44   &   \textbf{5}   &    11,025   &     1,936   &   \textbf{25}   &      54.4 &      22.8 &   \textbf{2.6} \\
Sunny-1    &        117 &        -38 &     \textbf{2} &      117   &       38   &   \textbf{2}   &    13,689   &     1,444   &    \textbf{4}   &       68.0 &      22.1 &   \textbf{1.2} \\
Sunny-2    &        113 &          -37 &     \textbf{2} &      113   &       38   &   \textbf{2}   &    12,769   &     1,444   &    \textbf{4}   &      66.1 &      22.2 &   \textbf{1.2} \\
\midrule
Mean       &          111.6 &      -36.1 &     \textbf{-0.5} &      111.6 &       36.3 &   \textbf{2.8} &    12,736.6 &     1,351.6 &   \textbf{10.5} &      63.9 &      20.7 &   \textbf{1.6} \\
\bottomrule
\multicolumn{13}{l}{B = Baseline \textbf{A}; S = Simplified algorithm \textbf{B}; O = Ours (mean aggr., IoU Threshold $\tau = 0.2$)}
\end{tabularx}

\label{tab:results_multi_camera_counting}
\end{table}

\section{Conclusion}
\label{sec:conclusions}
This paper presented a distributed artificial intelligence-based system that automatically counts the vehicles present in a parking lot using images taken by multiple smart cameras. Unlike most of the works in literature, we introduced a multi-camera approach that can estimate the number of cars present in the \textit{entire} parking area and not only in the single captured images. The main peculiarities of this approach are that all the computation is performed in a distributed manner at the edge of the network and that there is no need for any extra information of the monitored parking area, such as the location of the parking spaces, nor any geometric information about the position of the cameras in the parking lot. We modeled our system as a graph, where the nodes, i.e., the smart cameras, are responsible for estimating the number of cars present in their view and merging data from nearby devices that have an overlapping field of view. Our solution is simple but effective, combining a deep-learning technique with a distributed geometry-based approach. We evaluated our algorithm on the CNRPark-EXT dataset, which we specifically extended to show how we benefit from redundant information from different cameras while improving overall performance.

There are multiple lines of future development that can help improve the proposed system. Although our multi-camera algorithm is flexible, one limitation relies on computing the homographic matrix between images captured by cameras placed in completely different locations, such as facing each other. In this case, the two perspectives are totally different, and manual intervention is required to avoid the generation of an inaccurate geometric transformation.

Another interesting study that could be conducted is using the same approach in different scenarios, such as counting people in crowds using techniques based on density maps.

\section{Acknowledgments}
This work was partially supported by H2020 project AI4EU under GA 825619 and by H2020 project AI4media under GA 951911.

\bibliographystyle{apalike}  
\bibliography{references}

\begin{thebibliography}{}

\bibitem[Amato et~al., 2018]{amato2018wireless}
Amato, G., Bolettieri, P., Moroni, D., Carrara, F., Ciampi, L., Pieri, G.,
  Gennaro, C., Leone, G.~R., and Vairo, C. (2018).
\newblock A wireless smart camera network for parking monitoring.
\newblock In {\em {IEEE} Globecom Workshops, {GC} Wkshps 2018, Abu Dhabi,
  United Arab Emirates, December 9-13, 2018}, pages 1--6. {IEEE}.

\bibitem[Amato et~al., 2017]{amato2017deep}
Amato, G., Carrara, F., Falchi, F., Gennaro, C., Meghini, C., and Vairo, C.
  (2017).
\newblock Deep learning for decentralized parking lot occupancy detection.
\newblock {\em Expert Systems With Applications}, 72:327--334.

\bibitem[Amato et~al., 2016]{amato2016car}
Amato, G., Carrara, F., Falchi, F., Gennaro, C., and Vairo, C. (2016).
\newblock Car parking occupancy detection using smart camera networks and deep
  learning.
\newblock In {\em {IEEE} Symposium on Computers and Communication, {ISCC} 2016,
  Messina, Italy, June 27-30, 2016}, pages 1212--1217. {IEEE} Computer Society.

\bibitem[Amato et~al., 2019]{amato2019counting}
Amato, G., Ciampi, L., Falchi, F., and Gennaro, C. (2019).
\newblock Counting vehicles with deep learning in onboard {UAV} imagery.
\newblock In {\em 2019 {IEEE} Symposium on Computers and Communications, {ISCC}
  2019, Barcelona, Spain, June 29 - July 3, 2019}, pages 1--6. {IEEE}.

\bibitem[Arteta et~al., 2016]{arteta2016counting}
Arteta, C., Lempitsky, V.~S., and Zisserman, A. (2016).
\newblock Counting in the wild.
\newblock In Leibe, B., Matas, J., Sebe, N., and Welling, M., editors, {\em
  Computer Vision - {ECCV} 2016 - 14th European Conference, Amsterdam, The
  Netherlands, October 11-14, 2016, Proceedings, Part {VII}}, volume 9911 of
  {\em Lecture Notes in Computer Science}, pages 483--498. Springer.

\bibitem[Boominathan et~al., 2016]{boominathan2016crowdnet}
Boominathan, L., Kruthiventi, S. S.~S., and Babu, R.~V. (2016).
\newblock Crowdnet: {A} deep convolutional network for dense crowd counting.
\newblock In Hanjalic, A., Snoek, C., Worring, M., Bulterman, D. C.~A., Huet,
  B., Kelliher, A., Kompatsiaris, Y., and Li, J., editors, {\em Proceedings of
  the 2016 {ACM} Conference on Multimedia Conference, {MM} 2016, Amsterdam, The
  Netherlands, October 15-19, 2016}, pages 640--644. {ACM}.

\bibitem[Ciampi et~al., 2018]{ciampi2018counting}
Ciampi, L., Amato, G., Falchi, F., Gennaro, C., and Rabitti, F. (2018).
\newblock Counting vehicles with cameras.
\newblock In Bergamaschi, S., Noia, T.~D., and Maurino, A., editors, {\em
  Proceedings of the 26th Italian Symposium on Advanced Database Systems,
  Castellaneta Marina (Taranto), Italy, June 24-27, 2018}, volume 2161 of {\em
  {CEUR} Workshop Proceedings}. CEUR-WS.org.

\bibitem[Ciampi et~al., 2020]{DBLP:conf/ecai/CiampiSCGA20}
Ciampi, L., Santiago, C., Costeira, J.~P., Gennaro, C., and Amato, G. (2020).
\newblock Unsupervised vehicle counting via multiple camera domain adaptation.
\newblock In Saffiotti, A., Serafini, L., and Lukowicz, P., editors, {\em
  Proceedings of the First International Workshop on New Foundations for
  Human-Centered {AI} (NeHuAI) co-located with 24th European Conference on
  Artificial Intelligence {(ECAI} 2020), Santiago de Compostella, Spain,
  September 4, 2020}, volume 2659 of {\em {CEUR} Workshop Proceedings}, pages
  82--85. CEUR-WS.org.

\bibitem[Ciampi et~al., 2021]{DBLP:conf/visapp/CiampiSCGA21}
Ciampi, L., Santiago, C., Costeira, J.~P., Gennaro, C., and Amato, G. (2021).
\newblock Domain adaptation for traffic density estimation.
\newblock In Farinella, G.~M., Radeva, P., Braz, J., and Bouatouch, K.,
  editors, {\em Proceedings of the 16th International Joint Conference on
  Computer Vision, Imaging and Computer Graphics Theory and Applications,
  {VISIGRAPP} 2021, Volume 5: VISAPP, Online Streaming, February 8-10, 2021},
  pages 185--195. {SCITEPRESS}.

\bibitem[Dalal and Triggs, 2005]{dalal2005histograms}
Dalal, N. and Triggs, B. (2005).
\newblock Histograms of oriented gradients for human detection.
\newblock In {\em 2005 IEEE computer society conference on computer vision and
  pattern recognition (CVPR'05)}, volume~1, pages 886--893. Ieee.

\bibitem[Deng et~al., 2009]{deng2009imagenet}
Deng, J., Dong, W., Socher, R., Li, L., Li, K., and Li, F. (2009).
\newblock Imagenet: {A} large-scale hierarchical image database.
\newblock In {\em 2009 {IEEE} Computer Society Conference on Computer Vision
  and Pattern Recognition {(CVPR} 2009), 20-25 June 2009, Miami, Florida,
  {USA}}, pages 248--255. {IEEE} Computer Society.

\bibitem[Fischler and Bolles, 1981]{fischler1981random}
Fischler, M.~A. and Bolles, R.~C. (1981).
\newblock Random sample consensus: {A} paradigm for model fitting with
  applications to image analysis and automated cartography.
\newblock {\em Commun. {ACM}}, 24(6):381--395.

\bibitem[Gu et~al., 2009]{gu2009recognition}
Gu, C., Lim, J.~J., Arbelaez, P., and Malik, J. (2009).
\newblock Recognition using regions.
\newblock In {\em 2009 {IEEE} Computer Society Conference on Computer Vision
  and Pattern Recognition {(CVPR} 2009), 20-25 June 2009, Miami, Florida,
  {USA}}, pages 1030--1037. {IEEE} Computer Society.

\bibitem[He et~al., 2017]{he2017mask}
He, K., Gkioxari, G., Doll{\'{a}}r, P., and Girshick, R.~B. (2017).
\newblock Mask {R-CNN}.
\newblock In {\em {IEEE} International Conference on Computer Vision, {ICCV}
  2017, Venice, Italy, October 22-29, 2017}, pages 2980--2988. {IEEE} Computer
  Society.

\bibitem[He et~al., 2016]{he2016deep}
He, K., Zhang, X., Ren, S., and Sun, J. (2016).
\newblock Deep residual learning for image recognition.
\newblock In {\em 2016 {IEEE} Conference on Computer Vision and Pattern
  Recognition, {CVPR} 2016, Las Vegas, NV, USA, June 27-30, 2016}, pages
  770--778. {IEEE} Computer Society.

\bibitem[Lempitsky and Zisserman, 2010]{lempitsky2010learning}
Lempitsky, V.~S. and Zisserman, A. (2010).
\newblock Learning to count objects in images.
\newblock In Lafferty, J.~D., Williams, C. K.~I., Shawe{-}Taylor, J., Zemel,
  R.~S., and Culotta, A., editors, {\em Advances in Neural Information
  Processing Systems 23: 24th Annual Conference on Neural Information
  Processing Systems 2010. Proceedings of a meeting held 6-9 December 2010,
  Vancouver, British Columbia, Canada}, pages 1324--1332. Curran Associates,
  Inc.

\bibitem[Lin et~al., 2014]{lin2014microsoft}
Lin, T., Maire, M., Belongie, S.~J., Hays, J., Perona, P., Ramanan, D.,
  Doll{\'{a}}r, P., and Zitnick, C.~L. (2014).
\newblock Microsoft {COCO:} common objects in context.
\newblock In Fleet, D.~J., Pajdla, T., Schiele, B., and Tuytelaars, T.,
  editors, {\em Computer Vision - {ECCV} 2014 - 13th European Conference,
  Zurich, Switzerland, September 6-12, 2014, Proceedings, Part {V}}, volume
  8693 of {\em Lecture Notes in Computer Science}, pages 740--755. Springer.

\bibitem[Lowe, 1999]{lowe1999object}
Lowe, D.~G. (1999).
\newblock Object recognition from local scale-invariant features.
\newblock In {\em Proceedings of the International Conference on Computer
  Vision, Kerkyra, Corfu, Greece, September 20-25, 1999}, pages 1150--1157.
  {IEEE} Computer Society.

\bibitem[Lowe, 2004]{lowe2004distinctive}
Lowe, D.~G. (2004).
\newblock Distinctive image features from scale-invariant keypoints.
\newblock {\em Int. J. Comput. Vis.}, 60(2):91--110.

\bibitem[Nieto et~al., 2019]{nieto2018automatic}
Nieto, R.~M., Garc{\'{\i}}a{-}Mart{\'{\i}}n, {\'{A}}., Hauptmann, A.~G., and
  Mart{\'{\i}}nez, J.~M. (2019).
\newblock Automatic vacant parking places management system using multicamera
  vehicle detection.
\newblock {\em {IEEE} Transactions on Intelligent Transportation Systems},
  20(3):1069--1080.

\bibitem[O{\~{n}}oro{-}Rubio and L{\'{o}}pez{-}Sastre, 2016]{onoro2016towards}
O{\~{n}}oro{-}Rubio, D. and L{\'{o}}pez{-}Sastre, R.~J. (2016).
\newblock Towards perspective-free object counting with deep learning.
\newblock In Leibe, B., Matas, J., Sebe, N., and Welling, M., editors, {\em
  Computer Vision - {ECCV} 2016 - 14th European Conference, Amsterdam, The
  Netherlands, October 11-14, 2016, Proceedings, Part {VII}}, volume 9911 of
  {\em Lecture Notes in Computer Science}, pages 615--629. Springer.

\bibitem[Ren et~al., 2017]{Ren2015}
Ren, S., He, K., Girshick, R.~B., and Sun, J. (2017).
\newblock Faster {R-CNN:} towards real-time object detection with region
  proposal networks.
\newblock {\em {IEEE} Transactions on Pattern Analysis and Machine
  Intelligence.}, 39(6):1137--1149.

\bibitem[Sindagi and Patel, 2018]{sindagi2018survey}
Sindagi, V.~A. and Patel, V.~M. (2018).
\newblock A survey of recent advances in cnn-based single image crowd counting
  and density estimation.
\newblock {\em Pattern Recognition Letters}, 107:3--16.

\bibitem[V{\'{\i}}tek and Melnicuk, 2018]{vitek2018distributed}
V{\'{\i}}tek, S. and Melnicuk, P. (2018).
\newblock A distributed wireless camera system for the management of parking
  spaces.
\newblock {\em Sensors}, 18(1):69.

\bibitem[Xie et~al., 2018]{xie2018microscopy}
Xie, W., Noble, J.~A., and Zisserman, A. (2018).
\newblock Microscopy cell counting and detection with fully convolutional
  regression networks.
\newblock {\em Computer methods in biomechanics and biomedical engineering:
  Imaging \& Visualization}, 6(3):283--292.

\bibitem[Zhang et~al., 2017]{zhang2017understanding}
Zhang, S., Wu, G., Costeira, J.~P., and Moura, J. M.~F. (2017).
\newblock Understanding traffic density from large-scale web camera data.
\newblock In {\em 2017 {IEEE} Conference on Computer Vision and Pattern
  Recognition, {CVPR} 2017, Honolulu, HI, USA, July 21-26, 2017}, pages
  4264--4273. {IEEE} Computer Society.

\bibitem[Zhang et~al., 2016]{zhang2016single}
Zhang, Y., Zhou, D., Chen, S., Gao, S., and Ma, Y. (2016).
\newblock Single-image crowd counting via multi-column convolutional neural
  network.
\newblock In {\em 2016 {IEEE} Conference on Computer Vision and Pattern
  Recognition, {CVPR} 2016, Las Vegas, NV, USA, June 27-30, 2016}, pages
  589--597. {IEEE} Computer Society.

\end{thebibliography}

\end{document}